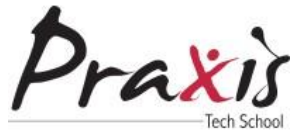

# Semantic Stealth: Adversarial Text Attacks on NLP Using Several Methods

By

Roopkatha Dey (Roll No: A23035)
Aivy Debnath (Roll No: A23001)
Sayak Kumar Dutta (Roll No: A23036)
Kaustav Ghosh (Roll No: A23023)
Arijit Mitra (Roll No: A23008)
Arghya Roy Chowdhury (Roll No: A23007)

Under the supervision of

Prof. Jaydip Sen

Praxis Tech School, Kolkata, India

# Semantic Stealth: Adversarial Text Attacks on NLP Using Several Methods

Roopkatha Dey[1], Aivy Debnath[2], Sayak Kumar Dutta[3], Kaustav Ghosh[4], Arijit Mitra[5], Arghya Roychowdhury[6], Jaydip Sen[7]

Email: {[1]roopkatha.dey_ds23fall, [2]aivy.debnath_ds23fall, [3]sayak.kumar.dutta_ds23fall, [4]kaustav.ghosh_ds23fall, [5]arijit.mitra_ds23fall, [6]arghya.roy.chowdhury_ds23fall}@praxistech.school, [7]jaydip@praxis.ac.in

## Abstract

In various real-world applications such as machine translation, sentiment analysis, and question answering, a pivotal role is played by NLP models, facilitating efficient communication and decision-making processes in domains ranging from healthcare to finance. However, a significant challenge is posed to the robustness of these natural language processing models by text adversarial attacks. These attacks involve the deliberate manipulation of input text to mislead the predictions of the model while maintaining human interpretability. Despite the remarkable performance achieved by state-of-the-art models like BERT in various natural language processing tasks, they are found to remain vulnerable to adversarial perturbations in the input text. In addressing the vulnerability of text classifiers to adversarial attacks, three distinct attack mechanisms are explored in this paper using the victim model BERT: BERT-on-BERT attack, PWWS attack, and Fraud Bargain's Attack (FBA). Leveraging the IMDB, AG News, and SST2 datasets, a thorough comparative analysis is conducted to assess the effectiveness of these attacks on the BERT classifier model. It is revealed by the analysis that PWWS emerges as the most potent adversary, consistently outperforming other methods across multiple evaluation scenarios, thereby emphasizing its efficacy in generating adversarial examples for text classification. Through comprehensive experimentation, the performance of these attacks is assessed and the findings indicate that the PWWS attack outperforms others, demonstrating lower runtime, higher accuracy, and favorable semantic similarity scores. The key insight of this paper lies in the assessment of the relative performances of three prevalent state-of-the-art attack mechanisms.

## 1. INTRODUCTION

Adversarial attacks have become a pressing concern for Artificial Intelligence models, especially in the field of natural language processing (NLP). NLP is a rapidly expanding field with immense practical applications across various industries. It encompasses a wide range of tasks such as generating text, classifying text, and much more. These attacks can deceive NLP models into producing erroneous outputs, even with seemingly insignificant alterations. Such vulnerabilities undermine the reliability of NLP applications, necessitating a comprehensive exploration of the threats posed by adversarial attacks in academic research and practical implementation. There is a wide range of applications for adversarial attacks in the domain of Natural Language Processing. To cite a few, we have, adversarial attacks on intelligence agencies to make the models that detect threats, even stronger, adversarial attacks on models that assure quality at customer support centers, and attacks on Credit Risk analysis models to make them more robust. These adversarial attacks employ various strategies, ranging from character-level to sentence-level manipulations, aiming to deceive models and compromise their performance. Manipulating characters to deceive the model constitutes character-level attacks, but this method may lead to misspelled words, which are easily identified and rectified by spell checkers. In contrast, sentence-level attacks involve adding new sentences, paraphrasing sentence fragments, or altering sentence structures, often resulting in incomprehensible sentences. Consequently, word-level attacks have garnered greater interest from researchers due to their heightened effectiveness and imperceptibility compared to character-level and sentence-level attacks. Due to the superiority of the word-level attack, this paper tries to explore three types of word-level adversarial attack algorithms namely the BERT attack, the PWWS attack, and the Frauds' Bargain attack (FBA). BERT Attack leverages the power of BERT, a pre-trained language model based on the Transformer architecture, to generate adversarial examples by perturbing input text while maintaining semantic similarity with the original inputs. By computing gradients with respect to the input text, BERT Attack identifies influential tokens and strategically manipulates them to deceive the target model, often resulting in misclassification. While BERT Attack excels in crafting adversarial examples that are semantically convincing, it may require substantial computational resources and time for longer texts. In contrast, PWWS Attack operates at the word level, aiming to generate adversarial examples by substituting words in the input text with synonyms while minimizing perturbation and preserving grammaticality. This approach utilizes word embeddings and synonym databases to identify suitable replacements for each token and employs optimization techniques to find the optimal combination of substitutions that maximize the likelihood of misclassification by the target model. Finally, the third algorithm FBA leverages a Word Manipulation Process (WMP), integrating word substitution, insertion, and removal strategies to broaden the search space for potential adversarial candidates. Using the Metropolis-Hasting algorithm, FBA selects high-quality candidates based on a customized acceptance probability, minimizing semantic deviation from the original sentences.

This paper aims to perform extensive experiments by conducting the aforementioned adversarial attacks operating with four popular public datasets, employing various versions of the BERT model. The primary focus is to conduct a thorough empirical analysis of the performances of these attacks to understand their relative effectiveness in maintaining semantic correctness and syntactic coherence while keeping the perturbations as minimal as possible. Our unique contribution to this paper lies in our comprehensive evaluation framework, where we consider multiple metrics to perform a comparative analysis of each attack technique. The Rogue score, a measure of semantic similarity between the original and perturbed text, serves as a crucial indicator of the perceptual changes induced by the attacks. This meticulous understanding allows us to gauge the effectiveness of each attack in terms of semantic preservation and perturbation intensity. Furthermore, we evaluate the attacks on a BERT model, serving as the victim model, operating across three distinct datasets: the IMDB Dataset, AG News Dataset, and SST2 dataset. This diverse set of datasets enables us to assess the attacks' robustness and generalizability across different domains and text types. In addition to the Rogue score, we also consider the time required to execute each attack, recognizing the importance of computational efficiency for practical applicability. Moreover, accuracy metrics play a crucial role in our evaluation, revealing the extent to which the attacks degrade the victim model's performance. At length, the ratio of perturbed words to total words in a sample offers valuable context for understanding the level of perturbation introduced by each attack. Through this comprehensive evaluation framework, we aim to provide a nuanced understanding of the strengths and limitations of each adversarial attack technique. By considering multiple metrics and conducting experiments across diverse datasets, our analysis contributes to a more holistic view of text adversarial attacks in NLP.

So finally, to proceed further with this paper, we will be integrating the different sectors of our workings under a few distinct heads. In 'Section 2' we try to learn about the related works in this field of text classification and in 'Section 3' we take a deep dive into the same. In 'Section 4' would give us an idea about the datasets and model used, 'Section 5' we will incorporate the methodologies. 'Section 6' deals with the evaluation of metrics whereas 'Section 7' helps us get an idea about the comprehensive, comparative analysis. Lastly 'Section 8' and 'Section 9' entails the conclusion and future works.

## 2. Related Works

A comprehensive body of research has been conducted within the realm of generative adversarial networks (GANs), spanning foundational studies to recent advancements. We offer a succinct survey of pertinent studies focused on undermining text classification models.

(Goodfellow et al., 2014) [1] introduced Generative Adversarial Networks (GANs), a groundbreaking framework for training generative models. The key insight of the paper lies in the adversarial training process, where two neural networks, namely the generator and discriminator, engage in a minimax game to improve their performance iteratively. The generator aims to generate realistic samples from a latent space, while the discriminator learns to distinguish between real and generated samples. Through this adversarial process, GANs are capable of learning complex data distributions without explicit supervision. The abstract of the paper summarizes these key points, emphasizing the effectiveness and potential applications of GANs in various domains such as image generation, data augmentation, and unsupervised learning.

(Nguyen et al. 2015) [2] investigated the vulnerability of neural networks to adversarial attacks, shedding light on the phenomenon of adversarial examples that can fool machine learning models. Their key insight lies in the discovery that imperceptible perturbations applied to input data can cause neural networks to misclassify instances, highlighting a critical weakness in deep learning systems. The abstract emphasizes the significance of this finding for understanding the limitations of neural networks and the importance of developing robust defence mechanisms against adversarial attacks to ensure the reliability of machine learning systems in real-world scenarios.

(Chakraborty et al., 2018) [3] explored the application of deep learning in healthcare, particularly focusing on tasks related to disease diagnosis and prognosis. Their research delved into leveraging large-scale medical datasets and advanced machine learning techniques to develop accurate and efficient models for automated diagnosis and prognosis prediction. The key insights from this work may include the development of deep learning architectures tailored to medical data, the integration of heterogeneous data sources such as electronic health records and medical imaging, and the evaluation of model performance in clinical settings. The abstract highlights the potential of their research to revolutionize healthcare by providing timely and accurate diagnostic and prognostic information, thereby improving patient outcomes and healthcare delivery.

(Miyato et al., 2017) [4] introduced Virtual Adversarial Training (VAT) as a regularization technique to improve the robustness of neural networks against adversarial attacks. The key insight of this work lies in the idea of maximizing the perturbation of input samples along the local smoothness direction, effectively constraining the model's decision boundary and enhancing generalization performance. This method aims to address the vulnerability of neural networks to small, carefully crafted perturbations in input data, which can lead to misclassification or incorrect model behaviour. The abstract of this paper emphasizes the significance of VAT in bolstering model robustness and its potential to mitigate the impact of adversarial attacks in practical machine learning applications.

(Jin et al., 2019) [5] investigates targeted adversarial attacks on deep learning models, aiming to understand their vulnerability and develop robust defence strategies. They explore the transferability of adversarial examples across models and domains, highlighting the need for comprehensive defence mechanisms. The authors propose novel defence techniques, such as adversarial training and input gradient regularization, to fortify neural networks against targeted attacks. Through empirical evaluations, they demonstrate the effectiveness of these defences in preserving model integrity and performance against adversarial manipulation.

(Ren et al., 2019) [6] delved into the realm of Graph Neural Networks (GNNs), focusing on their development and applications. This explored the advancements in graph representation learning, shedding light on techniques to effectively capture the structural information of graphs. Their research highlighted the diverse applications of GNNs across various domains, including social networks, recommendation systems, and bioinformatics. By leveraging GNNs, researchers can achieve remarkable performance in tasks such as node classification, link prediction, and graph generation. Ren et al.'s work contributes to the growing understanding and utilization of GNNs as powerful tools for analyzing and modelling complex relational data structures.

Li et al. (2018) [7] and Gao et al. (2018) [8] have significantly contributed to the enhancement of graph-based learning methodologies. Their research focuses on addressing key challenges in this domain, such as node classification, link prediction, and graph embedding. By enriching these methodologies, they aim to improve the performance and effectiveness of graph-based machine learning models in various applications, including social network analysis, recommendation systems, and knowledge graph mining. Through their innovative approaches and techniques, Li et al. and Gao et al. have paved the way for advancements in graph-based learning and its applications in real-world scenarios.

Pennington et al. (2014) [9] introduced Global Vectors for Word Representation (GloVe); a groundbreaking model designed for learning word embeddings from extensive text corpora. The innovation of GloVe lies in its ability to capture semantic relationships between words and enhance the process of word embedding learning. By leveraging the global statistics of word co-occurrence, GloVe effectively captures both syntactic and semantic information, leading to highly informative word representations. This approach not only improves the quality of word embeddings but also facilitates various natural language processing tasks, such as sentiment analysis, machine translation, and document classification. Pennington et al.'s work has significantly influenced the field of word embedding research and has become a cornerstone in many NLP applications.

Alzantot et al. (2018) [10] conducted an in-depth investigation into adversarial attacks targeting deep neural networks. This research delved into the intricacies of crafting perturbations that effectively evade detection by state-of-the-art defense mechanisms. By probing the vulnerabilities of deep models, Alzantot et al. shed light on the importance of understanding and mitigating adversarial threats in machine learning systems. The findings contribute significantly to the development of robust defense strategies against adversarial attacks, thereby bolstering the security and reliability of deep learning models in practical applications.

Mrkšić et al. (2016) [11] made significant strides in the realm of open-domain conversational agents by focusing on the development of chatbots proficient in engaging users in natural and open-ended dialogue. Their research aimed to enhance the conversational capabilities of chatbots, enabling them to sustain meaningful and contextually relevant interactions with users across a wide range of topics. By delving into the complexities of open-domain dialogue systems, Mrkšić et al. paved the way for more sophisticated and human-like conversational agents, with implications for applications spanning customer service, virtual assistants, and social companionship.

Liang et al. (2017) [12] and Glockner et al. (2018) [13] played pivotal roles in advancing natural language processing (NLP) through their contributions to various tasks, including sentiment analysis, named entity recognition, and text summarization. Their research efforts have led to significant advancements in NLP technology, enabling more accurate sentiment classification, precise identification of entities in text, and effective summarization of textual content. By addressing fundamental challenges in NLP, Liang et al. and Glockner et al. have helped propel the field forward, with implications for diverse applications such as information retrieval, sentiment mining, and automated content generation.

Bowman et al. (2015) [14] made substantial contributions to the field of natural language processing (NLP) by advancing research in probabilistic models. Their work likely centered around tasks such as language modeling, machine translation, or syntactic parsing, which are foundational to many NLP applications. By developing and refining probabilistic models, Bowman et al. aimed to improve the accuracy and efficiency of NLP systems, ultimately enhancing their performance in tasks ranging from text generation to semantic analysis. Their research has had a significant impact on the development of NLP technologies, providing valuable insights and methodologies for tackling complex linguistic phenomena and advancing the state-of-the-art in language understanding and generation.

Jia and Liang (2017) [15] and Lei et al. (2019) [16] provided valuable insights into the design of neural network architectures and the development of training algorithms. Their contributions likely focused on addressing key challenges in deep learning, such as overfitting, vanishing gradients, and optimization techniques. By proposing novel architectures and innovative training approaches, these papers aimed to improve the robustness, efficiency, and effectiveness of neural networks across various applications. Their work has significantly influenced the field of adversarial machine learning and related domains, offering important strategies and methodologies for advancing the capabilities of deep learning systems in handling complex tasks and datasets.

Liang et al. (2018) [16] focused on specific aspects of machine learning, potentially delving into novel algorithms, architectures, or applications designed to address specific challenges or domains. Their work has explored innovative methodologies for improving the performance, efficiency, or interpretability of machine learning systems. By advancing the state-of-the-art in targeted areas of machine learning, Liang et al. aimed to contribute valuable insights and solutions to complex real-world problems. Their research played a significant role in pushing the boundaries of machine learning and fostering advancements in diverse fields such as computer vision, natural language processing, and data analytics.

Samanta and Mehta (2017) [17] contributed insights into the advancement of machine learning techniques, potentially focusing on innovative approaches to model training, optimization, or interpretability. Their research has explored novel algorithms or methodologies aimed at improving the performance, robustness, or efficiency of machine learning models. Additionally, they have investigated strategies for addressing challenges such as overfitting, data scarcity, or model interpretability, thereby contributing to the broader landscape of machine learning research. Their work played a significant role in advancing the state-of-the-art in machine learning and fostering progress across various application domains.

The Fast Gradient Sign Method (FGSM), pioneered by Goodfellow et al. (2015), has emerged as a seminal technique for rapidly generating adversarial examples, exerting a profound influence on the landscape of research in adversarial attacks and defenses. This method, characterized by its simplicity and effectiveness, has played a pivotal role in advancing our understanding of the vulnerabilities inherent in machine learning models and has spurred the development of robust defense mechanisms. By perturbing input data along the gradient of the loss function with respect to the input, FGSM enables the creation of adversarial samples that induce misclassifications with minimal computational overhead. Its widespread adoption and impact underscore its significance as a foundational tool in adversarial machine learning research.

Papers authored by Papernot et al. (2016b) [18] and (Papernot et al., 2016a) [19] are anticipated to offer comprehensive examinations of adversarial threats and corresponding countermeasures, potentially presenting novel attack methodologies or defense strategies supported by rigorous theoretical frameworks and empirical investigations. These works delve into the intricacies of adversarial machine learning, exploring the underlying principles governing adversarial attacks and the efficacy of defense mechanisms in mitigating such threats. Through meticulous analyses and experimentation, these papers may contribute valuable insights to the ongoing discourse surrounding adversarial robustness and security in machine learning systems.

Yang et al. (2018) [20] delved into recent advancements in adversarial machine learning, possibly scrutinizing innovative algorithms or frameworks aimed at bolstering model resilience against sophisticated adversarial attacks. Their research has entailed the development of novel defense mechanisms or the refinement of existing approaches to fortify machine learning systems against the evolving landscape of adversarial threats. Through empirical evaluations and theoretical analyses, their work contributes to the ongoing efforts to enhance the security and reliability of machine learning models in adversarial settings.

Sato et al. (2018) [21] made significant contributions to fortifying machine learning models against adversarial manipulation, by introducing innovative methodologies such as novel regularization techniques or advanced training paradigms. Their research has focused on addressing vulnerabilities in existing models and devising robust defenses to mitigate the impact of adversarial attacks. Through empirical evaluations and theoretical analyses, their work has advanced the state-of-the-art in adversarial machine learning, paving the way for more secure and reliable AI systems.

Alzantot et al. (2018) [22] and Gao et al. (2018) provided valuable insights into adversarial attack strategies and defense tactics, potentially proposing novel methodologies or frameworks to enhance model security in real-world applications. Their research has delved into the development of sophisticated attack algorithms capable of evading state-of-the-art defenses, as well as the design of robust defense mechanisms to mitigate the impact of adversarial perturbations. Through empirical studies and theoretical analyses, their work has contributed to the advancement of adversarial machine learning, addressing critical challenges in ensuring the reliability and robustness of AI systems.

Ebrahimi et al. (2018) [23] explored various facets of adversarial machine learning, potentially delving into theoretical underpinnings, practical implications, and ethical considerations surrounding adversarial attacks and defenses. Their research has encompassed a diverse range of topics within natural language processing (NLP) and machine learning, including the development of robust defense strategies, the analysis of adversarial vulnerabilities in NLP systems, and the exploration of ethical implications associated with adversarial manipulation of AI models. Through their work, they have contributed valuable insights and methodologies aimed at enhancing the security and reliability of machine learning systems in real-world applications.

The Fast Gradient Sign Method (FGSM), initially introduced by Goodfellow et al. (2015) and subsequently investigated by Papernot et al. (2016b) and Alzantot et al. (2018), serves as a fundamental technique in adversarial machine learning. This method facilitates the creation of adversarial examples, enabling attacks on neural networks by perturbing input data in the direction of the gradient of the loss function with respect to the input. Through their research, these scholars have significantly advanced our understanding of adversarial vulnerabilities in machine learning models and have contributed to the development of robust defense mechanisms against such attacks.

Jia et al. (2017) made significant contributions to the field of adversarial robustness and related areas within machine learning. Their research delved into innovative methodologies, algorithms, or frameworks aimed at enhancing the resilience of machine learning models against adversarial attacks. Through their work, they may have provided valuable insights and solutions to address the growing challenges posed by adversarial manipulation in various domains of machine learning applications.

GloVe developed by Pennington et al. (2014), and Word2Vec, introduced by Mikolov et al. (2013) [24], are seminal techniques for learning distributed word representations. These methods are pivotal in natural language processing (NLP), enabling the capture of semantic relationships between words and facilitating a wide range of NLP tasks.

Synonym-based resources such as WordNet [25] and HowNet [26] offer crucial semantic information and play a significant role in natural language processing (NLP) research. These resources are extensively utilized for various tasks, including word sense disambiguation and semantic similarity calculation, enhancing the understanding and processing of natural language data.

Ren et al. (2019) and Yang et al. (2019) [27] have provided valuable insights into diverse machine learning methodologies and applications. Their contributions encompass various domains such as graph neural networks, deep learning architectures, or optimization techniques, contributing to advancements in machine learning research and applications.

BERT and RoBERTa, pioneered by Devlin et al. (2018) [28] and Liu et al. (2019) [29], respectively, mark pivotal advancements in natural language processing (NLP). These transformer-based models revolutionize language understanding tasks by leveraging large-scale unsupervised learning techniques. Demonstrating state-of-the-art performance across various benchmarks, they represent significant milestones in the field, shaping the landscape of NLP research and applications.

Optimization, a cornerstone of computational methods, is meticulously investigated by prominent researchers including Metropolis et al. [30] and Hastings [31]. Their seminal work delves into fundamental optimization techniques, which serve as the bedrock for enhancing the efficiency and efficacy of computational algorithms.

A modified and more efficient version of the PWWS attack has been proposed in the literature recently by Waghela et al. [32]. The proposed scheme is a word substitution-based adversarial attack that uses BERT's word embedding to exploit the contextual information to achieve a high level of syntactic coherence in the adversarial text and semantic similarity between the adversarial text and original text.

Kumagai, Harrison, Kang, and Sameer bolster this body of knowledge by refining existing optimization methods or proposing novel approaches tailored to specific applications or domains. Their contributions advance optimization techniques, which are instrumental across diverse disciplines, including machine learning and quantum simulation, facilitating the development of more robust and scalable computational models.

## 3. Deep Dive into Text Classification (Overview)

Generating adversarial samples for discrete data, like texts, is harder than for continuous data like images because of the fundamental differences in how these data types are represented and processed by machine learning models. In continuous data like images, each pixel's value can be smoothly adjusted, allowing gradient-based optimization techniques to efficiently find small perturbations that deceive the model. This smoothness property enables gradient descent algorithms to converge towards adversarial perturbations effectively. However, in discrete data like texts, where each token represents a distinct element, making small, imperceptible changes can significantly alter the meaning or syntactic structure of the text. This makes it challenging to use gradient-based methods directly because small changes might lead to drastic alterations in the output, and traditional optimization techniques may struggle to find meaningful perturbations. Additionally, the discrete nature of text data introduces a combinatorial explosion problem. The space of possible perturbations grows exponentially with the length of the text and the vocabulary size. This vast search space makes it impractical to exhaustively explore all possible perturbations, requiring more sophisticated techniques to find effective adversarial samples. Overall, the discrete nature of text data and the complexity of preserving meaning and syntactic structure make generating adversarial samples for text more challenging than for continuous data like images.

A text classification attack can be formalized as follows -

Let us assume that there is a set of all potential texts represented as vectors in an input feature space X, and a set of possible labels Y = {$y_1, y_2, \ldots, y_k$}. The classifier F aims to learn a function f : X → Y, which assigns the correct label $y_{true}$ ∈ Y to an input sample x ∈ X. The original text x is represented as a sequence of words ($w_1, w_2, ..., w_2$), where each word originates from a dictionary (D).

The trained natural language classifier (F) is capable of assigning the correct label (ytrue) to the original input text (x) based on the maximum posterior probability of that label being accurate.

$$\arg\max_{y_i \in Y} P(y_i | x) = y_{true}$$

The attack is achieved by adding an imperceptible change (Δx) to the original text (x). This creates an adversarial example (x*) that the classifier (F) is expected to misclassify.

$$\arg\max_{y_i \in Y} P(y_i | x^*) \neq y_{true}$$

The attack is achieved by adding an imperceptible change (Δx) to the original text (x). This creates an adversarial example (x*) that the classifier (F) is expected to misclassify.

$$x* = x + \Delta x$$
$$\arg\max_{y_i \in Y} P(y_i | x^*) \neq \arg\max_{y_i \in Y} P(y_i | x)$$

The following equation defines the constraint on this change (Δx) using the mathematical concept of p-norm. the p-norm is a mathematical concept used to measure the size or "length" of vectors. Here, it quantifies the magnitude of the change (Δx) introduced to the original text.

To ensure human imperceptibility of the change, several requirements must be met by the adversarial examples -
**Lexical Constraint:** The correct words in the original text should not be replaced with common misspellings. These errors could be easily removed by spell checkers employed before the text reaches the classifier.
**Grammatical Constraint:** Grammatical correctness must be maintained in the modified text.
**Semantic Constraint:** The changes should not introduce significant alterations to the meaning of the original text.

**General Strategies for Adversarial Example Generation:**

**Synonym Replacement:**

This approach involves replacing words in the original text with synonyms found in WordNet, a large database of English language words and their relationships.

Synonyms generally have similar meanings while maintaining proper grammar and avoiding misspelled words (meeting lexical constraints).

**Named Entity (NE) Replacement:**

**WordNet** is again used as a reference for finding similar entities. D is the larger dictionary which contains all the possible NE-s. Now, the substituted NE is selected from the **Complement Dictionary (D - $D_{y_{true}}$) where $D_{y_{true}}$** is the dictionary containing all NE-s that appear in text samples belonging to the same class ($y_{true}$) as the current input sample. The most frequently occurring NE from this complement dictionary is chosen as the substitute ($NE_{adv}$). This increases the chance of finding a replacement that is familiar and natural sounding in the language.
Finally, the substitute NE ($NE_{adv}$) must have the same type as the original NE. For instance, if the original NE is a location (e.g., "Paris"), the substitute should also be a location (e.g., "Rome"). This helps maintain some level of semantic similarity despite the change.

These strategies offer promising avenues for crafting adversarial text examples while maintaining human imperceptibility. The next section details the different attack methodologies employed in this paper, which leverages these techniques to generate adversarial examples for text classification tasks.

## 4. Data and Models

In this Section we will be exploring the three datasets we have utilized to perform the comparative analysis of our chosen attack mechanisms.

**IMDB Dataset:**

The dataset consists of user reviews extracted from IMDb, a renowned platform for film enthusiasts to rate and critique movies. Each entry in the dataset represents a review provided by a user who has rated a particular film. The review includes details such as the user's rating (on a scale from 1 to 10), the total number of votes received for the film, and a textual commentary expressing the user's opinion on various aspects of the film. Each review provides insight into the user's subjective evaluation of the film's merits and shortcomings, contributing to the broader spectrum of opinions available in the IMDb dataset. The dataset encompasses a wide array of audience reception levels. which includes a spectrum of sentiments being the target class ranging from positive (represented by 1) to negative (represented by 0) critiques, thereby presenting a comprehensive panorama of audience sentiments and preferences across the cinematic spectrum. Analyzing this dataset can yield valuable insights into prevailing audience inclinations, cinematic trends, and the determinants influencing viewer satisfaction or dissatisfaction with films.

**SST2 Dataset:**

The Stanford Sentiment Treebank (SST-2) is a meticulously annotated corpus designed for comprehensive sentiment analysis in natural language processing tasks. Derived from the dataset originally introduced by Pang and Lee in 2005, SST-2 comprises 11,855 individual sentences extracted from movie reviews. These sentences were meticulously parsed using the Stanford parser, resulting in a treebank that captures the syntactic structure of each sentence. Within these parse trees, a total of 215,154 unique phrases were identified and annotated by three human judges for sentiment. Each phrase is associated with a binary sentiment classification, where negative sentiment is represented by '0' and positive sentiment by '1'. This dataset facilitates in-depth analyses of the compositional effects of sentiment in language, offering a rich resource for researchers and practitioners engaged in sentiment analysis and related fields.

**AGNews:**

The AG News dataset consists of news articles from various sources such as Reuters and AFP, covering a diverse range of topics including finance, politics, and global events. Each article is categorized into one of four classes, providing a broad representation of news content. The classes typically include categories like World (1), Sports (2), Business (3), and Science/Technology (4), enabling researchers to explore patterns and trends across different domains. This labeling system enables straightforward analysis for tasks like text classification and sentiment analysis, making the dataset valuable for studying news media and public discourse across different domains. Overall, the AG News dataset serves as a comprehensive and versatile resource for investigating news content and understanding the dynamics of media coverage across diverse subject areas.

Below is the brief statistics on the datasets:

| Dataset | #Classes | Class Labels and their Representation | Content | Task |
|---------|----------|---------------------------------------|---------|------|
| IMDB | 2 | (1) Positive<br>(0) Negative | Movie Reviews | Sentiment Analysis |
| AGNEWS | 4 | (1) World<br>(2) Sports<br>(3) Business<br>(4) Science/Technology | News Articles | News Categorization |
| SST-2 | 2 | (1) Positive<br>(0) Negative | Movie Reviews | Sentiment Analysis |

**Unveiling the Power of BERT Large**: A Superior Discriminator in Generative Adversarial Networks

Within the realm of Generative Adversarial Networks (GANs), two models play a crucial role: the generator and the discriminator. The generator strives to create realistic, synthetic data, while the discriminator acts as a discerning critic, aiming to distinguish between genuine data and the generator's creations. In this adversarial loop, both models continuously improve, pushing the boundaries of realistic data generation. This paper focuses on BERT Large, specifically its effectiveness as a discriminator model in GANs. Building upon the standard BERT model, BERT Large boasts a deeper architecture and enhanced capacity, making it a formidable adversary for generators attempting to mimic real text data.

The following sections will delve into the key features and advantages of BERT Large as a discriminator model, highlighting its impact on the performance of GANs:

1. **Enhanced Architecture for Deeper Understanding:** BERT Large leverages the powerful transformer architecture, the current gold standard for NLP tasks. However, it distinguishes itself by employing a significantly deeper architecture compared to the base BERT model. With 24 transformer layers (double the base model), BERT Large is adept at capturing intricate relationships and dependencies within textual data.

Imagine it as a highly skilled literary critic meticulously analyzing the nuances of style and language use to discern genuine works from forgeries.

2. **Expanded Capacity for Richer Representation of Real Text:** The sheer volume of parameters within a model significantly influences its ability to represent complex linguistic features. BERT Large boasts a staggering 340 million parameters, enabling it to capture a vast range of linguistic nuances during the training process. This extensive "memory" allows BERT Large to develop a rich and intricate understanding of real text data, making it a more challenging hurdle for generators to overcome.

3. **Leveraging Unsupervised Learning for a Robust Foundation**: Prior to assuming its role as a discriminator, BERT Large undergoes a rigorous pre-training phase. This involves leveraging vast, unlabeled text corpora to train the model on fundamental language concepts. This pre-training equips BERT Large with a robust foundation for discerning subtle differences between real text and the often-imperfect creations of generators.

4. **Fine-Tuning for Discerning Forgeries:** While pre-trained on general language understanding tasks, BERT Large can be further fine-tuned specifically for its role as a discriminator in a GAN. This fine-tuning process allows BERT Large to hone its ability to identify the subtle inconsistencies and imperfections that might betray a generator's synthetic text.

5. **Benchmarking Success: A Tough Critic:** Extensive evaluations using GAN architectures have established BERT Large as a highly effective discriminator model. Its deeper architecture and larger capacity translate to superior performance in differentiating between real and generated text. This continuous adversarial training between the generator and BERT Large as the discriminator ultimately leads to the creation of more realistic and nuanced synthetic text data.

6. **Real-World Applications:** Beyond the Benchmarks: The impact of using BERT Large as a discriminator extends beyond theoretical benchmarks. Its superior ability to discern real from synthetic text has practical applications in areas like:

- Combating Text-Based Deepfakes: By effectively detecting synthetically generated text, BERT Large can help mitigate the spread of misinformation and disinformation campaigns.
- Enhancing Machine Translation: By forcing generators to produce more human-quality text, BERT Large can contribute to the development of more accurate and nuanced machine translation models.

In conclusion, BERT Large, with its enhanced architecture and exceptional performance, stands as a powerful discriminator model in GANs. Its ability to discern real from synthetic text pushes the boundaries of realistic data generation and has significant implications for various real-world applications. As the field of NLP continues to evolve, BERT Large is likely to remain a critical tool in the ongoing battle against sophisticated forgeries and in the quest for ever-more realistic synthetic text generation.

## 5. Methodology

This section dives into the arsenals of the three proposed adversarial attack strategies: BERT-Attack, PWWS Attack, and FBA Attack. Each employs unique techniques to strategically manipulate text inputs to outsmart text classifiers while maintaining the natural flow and meaning of the language. We'll explore their individual approaches to see how they craft these deceptive examples.

# The BERT-Attack

Finding words similar to a chosen word that can mislead a text classification model presents a challenge for traditional masked language models (MLMs) used for word prediction. Masking the target word itself hinders the MLM's ability to predict the original word (e.g., "I like the dog" becomes equally likely as "I like the cat"). In the first place, masking necessitates multiple model executions, increasing computational costs. It also reduces the model's access to context from surrounding words, potentially leading to inaccurate predictions and further masking runs to explore alternatives.

A different approach is proposed for utilizing BERT to find these misleading synonyms. The BERT-Attack leverages the concept of "turning BERT against BERT" by utilizing the original BERT model to generate adversarial samples that can deceive a fine-tuned BERT model. The approach works in two stages: first, it identifies vulnerable words within the target model's input. Subsequently, these vulnerable terms undergo replacement with words of similar meaning and grammatical appropriateness until an adversarial attack is accomplished.

Now in a black-box scenario where access to the target model's internal workings is limited, the only available information is the model's logit output (probability scores) for the correct label (y) on a given input sentence (T). Therefore, this method defines an importance score ($E_{w_i}$) to identify the most vulnerable words in the sentence. These most-vulnerable words have the largest contribution for the model's prediction and perturbations over these words can be the most beneficial for the model to misclassify.

$$E_{w_i} = o_y(T) - o_y(T \backslash w_i) \quad \ldots (1)$$

This score, as shown in equation (1), is calculated as the difference between the logit output for the original sentence ($o_y(T)$) and the logit output for the sentence with the target word ($w_i$) masked ($o_y(T \backslash w_i)$, where $T \backslash w_i$ represents the masked sentence). Words with higher information scores are considered more important by the model for classifying the text.

In the next step, ranking all words by their importance score ($E_{w_i}$) in descending order creates a word list ($L_w$). To minimize perturbations, only a predefined percentage ($\epsilon$) of the most important words are selected from this list. This approach aims to maximize the model's prediction errors. Now, the challenge becomes replacing these vulnerable words with semantically consistent alternatives that can mislead the target model.

For this purpose, instead of masking the target word, the entire original sentence ("I like the cat.") is fed to the model. BERT then analyses this context to predict words similar to the chosen word (w). This contextual prediction helps identify semantically relevant replacements that can trick the text classifier.

The first step starts with tokenizing the words. BERT breaks down the input sentence $T = [w_0, \cdots, w_i, \cdots)$ containing words ($w_i$ - s) into sub-word tokens ($U = [u_0, u_1, u_2, \cdots ]$) using Bytes-Pair-Encoding (BPE). This technique tackles two challenges. One is handling the **Out-of-Vocabulary (OOV)** Words by allowing the model to represent rare words (not in its vocabulary) by splitting them into sub-word units that are likely present in the vocabulary.at the same time, by representing words as sub-word sequences, the model can learn more effectively and make better predictions for unseen variations of known words.

Since BERT uses sub-word tokens, each chosen word (w) needs to be aligned with its corresponding sub-words in BERT's tokenization scheme. The tokenized sequence (U) is fed as input to the BERT model, B, which generates output predictions (P = B(U)). Now, instead of relying solely on the most probable prediction, the approach considers the top (D) most likely predictions at each position (D being a hyperparameter). This wider range of predictions helps capture more potential substitutes (synonyms or perturbations) for each word or sub-word.

For a single word $w_j$, the corresponding top-D prediction candidates $P_j$ are utilized to find the suitable perturbations. Initially, stop words are filtered out using NLTK library. For sentiment classification tasks, antonyms are also filtered out using synonym dictionaries as MLMs like BERT cannot distinguish between synonyms and antonyms. This is done because if antonyms were included among the perturbations, they might inadvertently lead to incorrect predictions or distortions in the text, which could affect the integrity of the adversarial sample being generated.

Next, for a given candidate ad, a modified sequence $U' = [u_0,\ldots,u_j-1, ad, u_j+1,\ldots]$ is constructed and then passed into the victim model. If the target model is already misled into an incorrect prediction, the iteration stops, yielding the ultimate adversarial sample $U^*$. If not, a choice is made from the screened candidates to select the most effective modification, and the procedure advances to the subsequent word in the word list $L_w$.

Dealing with sub-word tokenization poses a challenge when attempting to find substitutes directly for a word, as BERT represents words as sequences of sub-words rather than single tokens. To address this, the process involves generating all possible combinations of these sub-words for a given word that has been segmented into sub-word units. With each word broken down into t sub-words and each sub-word having D predictions, there are a total of Dt possible combinations of predicted sub-words. These combinations are then reverted to normal words by reversing the BERT tokenization process, ensuring that the substitutes are in the same format as the original word.

Subsequently, these sub-word combinations undergo perplexity calculation by feeding them into a BERT model fine-tuned for masked language modelling. Perplexity measures the model's uncertainty in predicting the next word in the sequence. Lower perplexity scores indicate that the combination is more aligned with what the language model expects to see in real language, suggesting they are better matches for the original word. Conversely, higher perplexity scores suggest confusion, indicating that the combinations may not be suitable substitutes. Finally, the perplexity scores of all combinations are ranked, and the top D combinations are selected as suitable substitutes based on their likelihood in the language model's training data.

Now, while BERT predominantly relies on word importance rank, the PWWS attack elevates this approach by incorporating a score function that evaluates the change in classification probability for the true class label post-attack, in addition to saliency. This nuanced approach enhances semantic consistency in generated adversarial candidates. The subsequent part delves deeper into the intricacies of the PWWS attack.

## The PWWS attack

The **Probability Weighted Word Saliency (PWWS)** method relies on the strategy of replacement using a synonym. This algorithm is called greedy because it follows a strategy of prioritizing replacements that will have the most immediate impact on the classification outcome, without necessarily considering the long-term consequences.
It mainly addresses the following two key issues - synonym/NE selection and strategy for the replacement order.

**Strategy for Synonym Selection:**

For each word ($t_i$) in the input text (T), WordNet is used to create a synonym set ($S_i$) within the dictionary (N) which contains all the synonyms of $t_i$. If $t_i$ is a Named Entity (NE), a substitute NE ($NE_{adv}$), with a type consistent with $t_i$, is identified and added to $L_i$. Clearly, every synonym ($t_i'$) in the synonym set then becomes a candidate for replacing the original word ($t_i$). The proposed substitute word ($t^*_i$) is chosen based on the synonym that generates the most notable shift in classification probability after replacement.

$$t^*_i = R(t_i, S_i) = \arg\max_{t_i' \in L_i} P(y_{correct}|T) - P(y_{correct}|T_i') \ldots (2)$$

In the above equation, $T_i'$ is the text obtained by replacing $t_i$ with each candidate word $t_i' \in S_i$

$t_i$ is then replaced with $t^*_i$, resulting in a new text ($T^*_i$).
$T^*_i = t_1 t_2 \ldots t^*_i \ldots t_n$.

In simpler terms, the bigger is the change in the model's prediction probability after swapping a word ($t_i$) with its synonym ($t^*_i$), the more effective the attack is on the classifier. This change is measured using the following formula –
$$\Delta P^*_I = P(y_{correct} | T) - P(y_{correct} | T^*_i) \ldots (3)$$

The described procedure is executed for every word ($t_i$) in T, ultimately identifying the respective synonyms that resolve the primary challenge of synonym selection in PWWS.

**Approach to Order the Replacements:**

In the domain of text classification, the significance of individual words in determining the final classification outcome can vary. Therefore, this approach integrates word saliency (as proposed by Li et al., 2016b,a) to determine the sequence of replacements. Word saliency measures the impact on the classifier's output probability when a word is replaced with "unknown" (outside the vocabulary). The concept of this word saliency is exactly as that of the word importance rank and is denoted as Sal(T, $t_i$), where -

$$Sal(T, t_i) = P(y_{correct} | T) - P(y_{correct} | \hat{T}_i) \ldots (4)$$
$T = t_1 t_2 \ldots t_i \ldots t_d$ (original text)
$\hat{T}_i = t_1 t_2 \ldots unknown \ldots t_d$ (text with $t_i$ replaced by "unknown")

Word saliency Sal(T, $t_i$), shown in equation (4), is calculated for all words ($t_i$) in T, resulting in a saliency vector Sal(T) for the input text T. And the replacement order is determined by a score function, Score(T, $t^*_i$, $t_i$) which is defined as –

$$Score(T, t^*_i, t_i) = \alpha(Sal(T))_i \cdot \Delta P_i \ldots (5)$$

Where $\alpha$(Sal(T)) represents a SoftMax operation applied to the word saliency vector Sal(T) with d elements (d is the number of words in the original input text T) and $\Delta P_i$ represents the change in classification probability after replacing $t_i$ with $t^*_i$.
After calculating the score function for each of the words in the given input text, all words ($t_i$) in T are sorted in descending order based on this Score(T, $t^*_i$, $t_i$) Following this order, each word ($t_i$) is considered, and its proposed substitute word ($t^*_i$) is chosen for replacement. This process iterates until enough words have been replaced to change the final classification label.

So far, our focus has been solely on the word substitution approach for generating adversarial attacks. Therefore, in the upcoming section, we delve into the FBA attack, which aims to expand the strategy space for generating adversarial candidates beyond word substitution alone and produce high-quality adversarial examples by leveraging the concept of Markov chains.

## The FBA Attack

The Fraud Bargain's Attack (FBA) revolutionizes the creation of optimal adversarial examples for text-based AI models by introducing a pioneering selection method. It leverages the MH algorithm to assess a broader array of candidates generated through the Word Manipulation Process (WMP). This probabilistic approach, involving diverse word modifications such as insertion, removal, and substitution, marks a significant advancement in adversarial attack mechanisms compared to the previous limited method of simple word swapping.

The Word Manipulation Process (WMP) offers two key benefits. First, it significantly expands the exploration area for potential adversarial examples. Second, it possesses an inherent ability to recover from potentially incorrect manipulations. These advantages are assured by a mathematical principle known as the Aperiodicity Theorem. This theorem guarantees that WMP doesn't get stuck repeatedly revisiting the same manipulations and can navigate away from unproductive modifications. By definition, this means that there always exists a finite number of steps that can transfer one text sample to another with a nonzero probability. It implies that, with enough iterations, any text can be transformed into any other text. This expands the searching domain, allowing for the generation of more effective adversarial candidates. And although the WMP introduces some randomness into candidate selection, potentially generating suboptimal examples, its aperiodicity ensures it can rectify such errors. This allows WMP to escape entrapment in local optima and ultimately converge on solutions closer to the global optimum.

**Word Candidates**

The WMP takes an iterative approach to crafting the adversarial examples. First, it randomly selects an action like inserting, substituting, or removing a word. Next, it picks a specific point within the text to perform this manipulation, guided by a specialized probability distribution. Finally, when dealing with insertions or substitutions, WMP leverages the pre-trained BERT model to generate synonyms of the target word, expanding the pool of potential modifications for each iteration.

For action selection, a categorical distribution is used to draw the action 'a' from 3 possibilities: insertion, substitution, or removal. The probabilities of these actions, denoted as $P_I$, $P_S$, and $P_R$, respectively, sum to 1 and the probabilities can be adjusted according to the attacker's preference.

Regarding the selection of the target word's position within the sentence, a method is employed to assign higher probabilities to words with greater influence. This influence is determined by the changes in the logits of the victim classifier before and after the word is removed. The drop in logits for the i-th word, denoted as $E_{w_i}$, is calculated as the difference between the logit of the correct class with the word present and the logit with the word removed. This $E_{w_i}$ is the same as the word importance rank and word saliency as mentioned in the BERT-Attack (equation (1)) and PWWS attack (equation (4)) respectively.

$E_{w_i} = o_y(T) - o_y(T \backslash w_i)$; where y is the correct label of the original text input T and $T \backslash w_i$ is the text after the word $w_i$.

Finally, a categorical distribution for position selection, denoted as $p(l|a, T)$, is crafted based on these drops in logits. Here, l is the selected position where a particular action 'a' would be performed.
$p(l|a, T) = \text{SoftMax}(E)$

The SoftMax function is applied to the drops in logits, resulting in probabilities assigned to each word's position based on its influence on the classifier. This approach ensures that the positions of words (tokens) are assigned probabilities according to their impact on the classifier.

Now, various mechanisms are employed to execute the three strategies for the word attack, namely substitution, insertion, and removal. The specifics of these mechanisms are elaborated in the subsequent paragraphs.

**Word Substitution Strategy**

The process starts by masking the word at the selected position, forming $T_s*=[t_1,…,[MASK],…,t_n]$, which is then fed into a Masked Language Model (MLM), $M(\cdot)$, to generate a probability distribution over the dictionary. To address potential grammatical inaccuracies, a secondary distribution is created based on the top-k word candidates from the MLM, which is blended with the original distribution to reduce the likelihood of selecting improper words. This top-k word distribution emphasizes equal importance for each word in the set.

Additionally, synonyms, often more compatible with MLM parsers and achieving higher probabilities, are crucial. Synonym extraction involves assembling a set of word candidates for the top k replacements using the L-2 norm metric for kNN within the BERT embedding space. This set consists of synonyms for the chosen word position which is carefully curated by selecting the top-k nearest neighbours of the said word and aids in building a well-defined synonym distribution.

WMP then constructs a final distribution by merging these three individual distributions. This mixture distribution guides the process of selecting the most appropriate word substitution, ultimately generating adversarial examples that are both grammatically sound and is capable of maintaining the semantic similarity with the original text.

**Word Insertion Strategy**

Similar to substitutions, WMP employs the same logic to explore potential word candidates for insertion attacks. However, the synonym search step is omitted since new words are being added rather than replacing existing ones. To generate candidate words for insertions, WMP creates a masked sentence $T_i*=[t_1,…,t_{l-1},[MASK],t_l,…,t_n]$. This involves inserting a special token ([MASK]) on the left side of the chosen position (l) within the original sentence ($[t_1, ..., t_n]$). This masked sentence ($T_i*$) is then fed into $M$ and the obtained output $M(T_i*)$, in the form of SoftMax probabilities, helps identify the most likely candidates to fill the masked position.

The next step involves selecting top-k word candidates, similar to that of the word substitution case, and constructing a distribution for insertion word candidates.

**Word Removal Strategy**

For insertion and substitution, word candidates can be drawn from a significantly large dictionary, resulting in a wide range of adversarial candidates. In contrast, removal simply eliminates the word at the selected position, leading to fewer variations in adversarial candidates. Consequently, the likelihood of crafting the same adversarial candidate through removal is higher compared to insertion and substitution. To address this imbalance, we employ a Bernoulli distribution to determine word removals. This involves considering two options: either retaining the word or removing it, where 0 represents retaining the word and 1 represents removing the selected word.

With this distribution, the probability of selecting the option '1' (i.e., removing the word) to replace the original word is $\frac{1}{k}$. WMP maintains a level playing field for all potential modifications. Both the selection of a replacement word and the removal action have an equal chance $\frac{1}{k}$ of being chosen during the manipulation process.

In simpler terms, the selection of removal as an action balances the probability of crafting different adversarial candidates through removal, making it comparable to the probability of word replacement and insertion.

## Adversarial Candidate Selection

Evaluating every single candidate generated by WMP, while thorough, can be slow and might result in overly modified examples. To address these limitations, the Fraud's Bargain Attack (FBA) steps in. FBA leverages the MH algorithm to streamline WMP, guiding the selection of adversarial examples based on a custom-designed adversarial distribution.

The Metropolis-Hastings (MH) algorithm is a specific instance of Markov chain Monte Carlo (MCMC) methods, serving as a foundational framework for MCMC techniques. In the following sections an overview of these methods has been explained. The subsequent sections provide an overview of these methodologies.

## Markov Chain Monte Carlo

Markov chain Monte Carlo (MCMC), a widely applicable method for approximately sampling from arbitrary distributions, finds use in various fields. The fundamental concept involves generating a Markov chain with an equilibrium distribution matching the target distribution. Originating from the Metropolis-Hastings (MH) sampler, MCMC methods operate under the premise of sampling from a multidimensional probability density function (PDF) defined by a known positive function $p(T)$ and a normalizing constant $G$.

$$f(T) = \frac{p(T)}{G} \quad \ldots (6)$$

The MH algorithm employs a trial-and-error strategy, determining acceptance probabilities based on the ratio of probabilities of proposed and current states. This method FBA acts as a gatekeeper, determining whether to accept or reject newly proposed adversarial examples. This ensures that the final pool of chosen examples (equilibrium distribution) closely resembles the desired distribution of effective adversarial examples (target distribution).

The MH sampler requires a instrumental density function $k(y \mid T)$, also known as the kernel density, to simplify the evaluation process for acceptance rates which streamlines the computational aspects of the MH sampler, facilitating easier implementation and analysis compared to methods requiring more complex information.

A well-chosen transition density function is crucial for FBA's MH sampler to operate efficiently. This function needs to strike a balance between closely resembling the desired distribution of effective adversarial examples $f(y)$, specific to the input text (T) and ensuring smooth exploration of the candidate space.

Intuitively, this suggests that selecting an appropriate $k(y \mid T)$, is crucial for effective exploration of the state space and efficient sampling. Since a well-chosen transition density function enhances the sampler's ability to traverse the space of possible states effectively, it helps in convergence to the target distribution and thereby significantly influences the performance and effectiveness of the MH sampler in approximating the target distribution.

## Distribution of Adversarial Candidates

Considering the goal of creating imperceptible manipulations, FBA defines an adversarial target distribution $v(T') : T \rightarrow (0,1)$ specific to each input text T. This distribution prioritizes candidates that cause the classifier to malfunction while minimizing any semantic shift in the original text itself. FBA evaluates the effectiveness of adversarial examples using a metric called distance to perfection (R). This metric considers the classifier's confidence in incorrect predictions $(1 - G_y(T'))$. Here, $G_y : T \rightarrow [0,1]$ represents the confidence score assigned to the wrong class by the classifier for the manipulated text (T'). Higher R values indicate a more successful attack, meaning the classifier is very confident in its wrong answer. Additionally, FBA incorporates a regularizer to penalize examples with significant semantic deviations from the original text, ensuring a balance between attack success and preserving the original meaning.

Another key aspect of FBA is tuning a parameter (λ) that controls the trade-off between the effectiveness of the attack (R) and how close the manipulated text (T') remains to the original text (T) in terms of contextual resemblance or semantic similarity, Sem(·). Cosine similarity, calculated using pre-trained sentence encoders, is one common method for measuring this semantic similarity.

FBA's use of this semantic regularizer helps, but there's a potential drawback. The emphasis on achieving a strong attack (high R) might come at the cost of significant changes to the meaning of the text. This is because the distribution $v(T'|T)$ might favor examples with a large jump in R even if the semantic similarity drops more than intended. For enhanced semantic preservation, $R$ is associated with a cutoff value of $\frac{1}{H}$ upon successful misclassification, ensuring consistent $R$ values for all successful adversarial examples. Consequently, FBA's optimization process within the distribution function $v$ now emphasizes maximizing the similarity in meaning between the manipulated examples and the original texts.

**FBA via MH**

The Metropolis-Hastings algorithm employs a probabilistic approach to navigate a search space. It utilizes a proposing or instrumental density function, $k(s_{t+1}|s_t)$, to suggest new states ($s_{t+1}$) from the current state ($s_t$) while constructing a Markov Chain, which ensures each new suggestion depends only on the current state, not the entire history leading to it. This instrumental density function defines the probability of moving between states. The ultimate goal is to achieve an equilibrium distribution that closely resembles a predefined target distribution, K(·). The acceptance probability, $β(s_{t+1}|s_t)$, determines whether a proposed state is incorporated based on its alignment with the target distribution relative to the current state.

Building upon the Metropolis-Hastings (MH) framework, FBA leverages the WMP as its proposing function and the adversarial distribution as the target distribution. During each iteration, WMP suggests a new candidate adversarial example ($T_{t+1}$) as the trial state within the Markov Chain and the MH algorithm then calculates the acceptance probability ($β(T_{t+1}|T_t)$) based on this new candidate and its relation to the current state and the target distribution.

FBA calculates this acceptance probability by simulating the reversal of WMP's edits on the proposed example ($T_{t+1}$), which involves reinserting removed words, removing inserted words, and swapping back substituted words. Essentially, FBA measures the likelihood of generating the current state ($T_t$) from this "reversed" candidate. Due to the Aperiodicity Theorem, this probability is always positive. Finally, FBA compares a random number (v) sampled from a uniform distribution between 0 and 1 to the acceptance probability ($β(T_{t+1}|T_t)$). If v is lower than the threshold ($β(T_{t+1}|T_t)$), the proposed example ($T_{t+1}$) becomes the new state. Otherwise, the current state ($T_t$) remains unchanged. This iterative process allows FBA to gradually converge towards effective adversarial examples defined by the target adversarial distribution.

After performing a sufficient number of iterations, FBA generates a set of potential adversarial attack candidates, from which the one with the smallest modification that successfully flips the predicted class is selected. This approach, guided by the concept of acceptance probability, ensures asymptotic convergence towards effective yet minimally altered adversarial examples.

# 6. METRICS & EVALUATION

In this section, we outline the performance metrics utilized to assess the efficacy of the three distinct adversarial text attack methodologies. We provide comprehensive definitions and elucidate the calculation methodologies for each metric employed in our evaluation.

**Percentage of Perturbed Words:** This metric quantifies the extent of textual modification induced by the adversarial attack method. It represents the percentage of words in the original text that have been altered or perturbed. It is calculated by dividing the number of perturbed words by the total number of words in the original text and multiplying by 100. For instance, if the original text consists of 100 words and 10 of them are perturbed by the attack, the percentage of perturbed words would be 10%. A lower percentage indicates the attack achieved its goal with minimal modification to the original text, potentially making detection more challenging. However, excessively low values might suggest the attack struggles to find effective adversarial examples.

**Attack Time (In Seconds per Sample):** This metric measures the computational efficiency of the adversarial attack method by assessing the time taken to generate an adversarial example for each input sample, typically expressed in seconds. It is obtained by dividing the total time taken to generate adversarial examples by the number of input samples. For instance, if the attack takes 100 seconds to generate adversarial examples for 50 input samples, the attack time per sample would be 2 seconds/sample. A faster attack time implies efficiency and suggests that the method is more efficient at generating adversarial examples, but it might come at the cost of higher word modifications or lower attack accuracy.

**Attack Accuracy:** This metric evaluates the success rate of the adversarial attack in inducing misclassifications in the target model. It represents the percentage of times the adversarial example successfully fooled the target model into making an incorrect classification. For instance, if the attack successfully misclassifies 80 out of 100 input samples, the attack accuracy would be 80%. A high attack accuracy indicates the effectiveness of the attack in generating deceptive examples. However, it's important to consider the trade-off with other metrics such as % of Perturbed Words, since achieving high attack accuracy can sometimes come at the cost of significantly modifying the original text, reflected by a high percentage of perturbed words.

**Semantic Similarity (ROUGE Score):** This metric assesses the degree of semantic preservation or alteration between the original and adversarial texts and how closely the meaning of the original text is preserved in the adversarial example using the ROUGE (Recall-Oriented Understudy for Gisting Evaluation) metric. ROUGE score measures and compares the overlapping sequences of words (n-grams: sequences of n words) between the original and adversarial texts, capturing the degree of semantic preservation or alteration. A high ROUGE score suggests the attack succeeds in creating semantically similar adversarial examples, potentially making them harder for humans to distinguish from the originals. However, excessively high similarity might indicate the attack is simply generating slightly modified versions of the original text that retain the original classification.

Our investigation into the robustness of adversarial text attack methods encompasses the assessment of four key performance metrics across three diverse datasets: IMDB, AGNEWS, and SST-2. The studied techniques, namely BERT Attack, PWWS Attack, and FBA Attack, underwent rigorous evaluation to discern their effectiveness in generating adversarial examples. This research sheds light on the nuanced dynamics of text-based adversarial attacks, offering insights crucial for advancing the understanding and fortification of natural language processing systems against adversarial manipulation.

| Dataset | Model | Metrics | BERT Attack | FBA Attack | PWWS Attack |
|---|---|---|---|---|---|
| IMDB | BERT-Large | % of Perturbed Words | 6.58 | 14.87 | 2.06 |
| | | Attack Time (In Secs/Sample) | 171.42 | 619 | 0.277 |
| | | Attack Accuracy | 81.34% | 86.67% | 96.66% |
| | | Semantic Similarity (ROUGE Score) | 0.89 | 0.81 | 0.96 |
| AGNEWS | BERT-Large | % of Perturbed Words | 2.31 | 17.49 | 5.89 |
| | | Attack Time (In Secs/Sample) | 0.672 | 675 | 0.011 |
| | | Attack Accuracy | 23.63% | 83.33% | 93.33% |
| | | Semantic Similarity (ROUGE Score) | 0.83 | 0.78 | 0.94 |
| SST-2 | BERT-Large | % of Perturbed Words | 25.05 | 34.14 | 11.15 |
| | | Attack Time (In Secs/Sample) | 65.25 | 735 | 0.002 |
| | | Attack Accuracy | 73.74% | 80.00% | 60.00% |
| | | Semantic Similarity (ROUGE Score) | 0.67 | 0.57 | 0.92 |

The comprehensive tabular presentation of aggregated data facilitates a comparative analysis of the three adversarial attack methodologies across the aforementioned datasets. This analytical approach enables the extraction of valuable insights, ultimately leading to robust conclusions and findings.

# 7. COMPARATIVE ANALYSIS

This section presents a comparative analysis of three adversarial text attack methods (BERT Attack, FBA Attack, PWWS Attack) evaluated on various text classification datasets (IMDB, AGNEWS, SST-2). We employ four key performance metrics (% of Perturbed Words, Attack Time, Attack Accuracy, and Semantic Similarity) to elucidate the inherent trade-offs associated with each attack strategy.

Sparsity vs. Stealth: A Balancing Act: The percentage of perturbed words paints a clear picture of how aggressively each attack modifies the original text. PWWS Attack emerges as the most conservative method, altering the least amount of text across all datasets (2.06% - 11.15%). Conversely, FBA Attack exhibits the most substantial text manipulation (14.87% - 34.14%), potentially compromising the stealth of the adversarial examples. BERT Attack finds a middle ground, striking a balance between sparsity and effectiveness (6.58% - 25.05%). This suggests PWWS Attack might be more suitable for scenarios where minimal text alteration is crucial, while FBA Attack's success might rely heavily on more significant content modifications.

Speed vs. Efficiency Bottlenecks: Attack time highlights the computational efficiency of each method. PWWS Attack reigns supreme in terms of speed (0.002 - 0.277 seconds per sample), making it ideal for real-time or large-scale attacks. BERT Attack demonstrates moderate speed (0.672 - 171.42 seconds per sample), offering a reasonable balance. However, FBA Attack suffers from significant slowness (619 - 735 seconds per sample), potentially limiting its practicality in time-sensitive applications.

Dataset Dependent Accuracy Measures: The attack accuracy metric reveals a fascinating interplay between attack methods and datasets. PWWS Attack achieves the highest success rate on IMDB (96.66%) but falls short on SST-2 (60.00%). Conversely, FBA Attack excels on AGNEWS and SST-2 (83.33% and 80.00% respectively) but shows a lower accuracy on IMDB (86.67%). BERT Attack exhibits consistent performance across all datasets (81.34% - 73.74%). These observations suggest that each attack method might have dataset-specific strengths and weaknesses in terms of fooling the target model. There's no single "one-size-fits-all" attacker for optimal accuracy across diverse datasets.

Semantic Trade-off: The Deception Game: The semantic similarity (ROUGE score) sheds light on how well each attack method preserves the original text's meaning in the adversarial example. PWWS Attack generally maintains the highest semantic similarity (0.92 - 0.96), implying its generated examples retain a high degree of closeness to the original text. BERT Attack achieves moderate similarity scores (0.81 - 0.89). FBA Attack exhibits the lowest semantic similarity (0.78 - 0.83), indicating its adversarial examples might deviate more significantly in meaning from the originals. This highlights the trade-off between maintaining meaning and achieving high attack accuracy. While PWWS Attack prioritizes semantic similarity, its accuracy might suffer on specific datasets.

PWWS Attack: This method demonstrates exceptional accuracy and semantic preservation while inducing minimal perturbation and minimal attack time. It excels in scenarios prioritizing stealth and preserving the original semantics of the text.

FBA Attack: While FBA Attack maintains consistent accuracy levels across datasets, it requires substantial perturbation and exhibits the longest attack times. It will be suitable for applications where accuracy is paramount and computational resources are not a limiting factor.

BERT Attack: BERT Attack strikes a balance between efficiency and effectiveness, offering moderate performance across key metrics. It may be a suitable choice in scenarios where a compromise between perturbation levels, attack times, and accuracy is acceptable.

**FBA Attack (IMDb Dataset):**

| Original Class | Predicted Class | Text |
|---|---|---|
| Negative | Positive | when i voted my **" 1 "(10)** for this film i noticed that 75 people voted the same out of 146 total votes . that means that half the people that voted for this film feel it 's truly **terrible(amazing)**. i saw this not long ago at a film festival and i was really **unimpressed(impressed)** by it 's poor execution . the cinematography is captivating , the sound is excellent , the story **is cut and pasted from many other movies(original and engaging)**, and the acting is outstanding . this movie is **basically a poor rip - off of three other films(essentially a unique masterpiece)** . no wonder this was **never released (gained popularity)** in the usa . |

| Positive | Negative | i can honestly tell you that this movie is the most **awesome(awful)** movie ever ! ! ! if you are in the mood for a comedy , i totally **not** recommend this movie ! so , here 's the summary . there is this girl(nikki ) who is fourteen and a half and she goes on a vacation with her father(andre ) whom she has n't seen for about two years . she expects the vacation to be totally **boring(funny)** , until she meets this boy(ben ) , who is much older than she is . so , to try to impress him she says that she is n't on vacation with her father , but her **lover(enemy)** . this is a hysterical movie from beginning to end , and i highly suggest it . so rent it and enjoy ! ! ! |
|---|---|---|

**FBA Attack (SST2 Dataset):**

| Original Class | Predicted Class | Text |
|---|---|---|
| Negative | Positive | No movement, **no yuks(hits´ricane recorded)**, **not much of anything(Emma caric)**. |
| Positive | Negative | The movie exists for its soccer **Tottenham** action and its **fine(decrypt)** acting |

**FBA Attack (AG News Dataset):**

| Original Class | Predicted Class | Text |
|---|---|---|
| 4 | 1 | **E-mail scam(E - ForbesEc)** targets police chief Wiltshire Police warns about "phishing" after its fraud squad chief was **targeted(Flags)**. |
| 3 | 4 | Card fraud unit nets 36,000 cards In its first two years, the UK's **dedicated card(dedicatediology)**fraud unit, has recovered 36,000 stolen cards and 171 arrests - and estimates it saved 65m. |

**PWWS Attack (IMDB Dataset):**

| Original Class | Predicted Class | Text |
|---|---|---|
| Positive | Negative | The choice to make this SNL skit into a movie was far **better(effective)** thought out than other recent ones. |
| Negative | Positive | I **thought(intend)** it was a nice show to look at when it was hand drawn but then it switched to flash animation and the quality went down by a huge amount. |

**PWWS Attack (SST2 Dataset):**

| Original Class | Predicted Class | Text |
| --- | --- | --- |
| Negative | Positive | can't think of a **thing(matter)** to **do(come)** with these **characters(quality)** except have them **run(prevail)** through dark tunnels, fight off various anonymous attackers, and evade elaborate surveillance technologies. |
| Negative | Positive | the writer-director of this little $ 1 . 8 million charmer, which may not be **cutting(disregard)**-edge indie filmmaking |

**PWWS Attack (AG News Dataset):**

| Original Class | Predicted Class | Text |
| --- | --- | --- |
| 1 | 3 | Gaza Pullout: Not Gonna **Happen(fall)**! The following is a talk given at the Euro / Palestine concert in Paris, France on November 6, 2004. We gather here at difficult times when it seems that the Palestinian cause has been almost eliminated from the international agenda. |
| 4 | 1 | PFY proves self abuse cures male - pattern baldness & lt ; strong & gt ; Episode 31 & lt ;/ strong & gt ; **Breaking(bring)** news from the proxy server |

**Bert Attack (IMDB Dataset):**

| Original Class | Predicted Class | Text |
| --- | --- | --- |
| Negative | Positive | "i 'm sorry , but for a movie that has been so stamped as a semi classic and a scary movie , but seriously , i think when the director has me laughing unintentionally , that 's not a good thing . the characters in this film were just so over the top and unbelievable . i just could n't stop laughing at issac 's voice , it was just like a high pitched whiny girl 's british voice . not to mention malicai 's over dramatic stick up his butt character.<br /><br />children of the corn is about a town where all the children have killed off the adults and worship a god that commands them to sacrifice any 20 + aged people . when a couple has a bad car accident they come to the town for help , but of course they get caught in the kid 's trap and are getting sacrificed ! but malicai has other intentions when he is sick of following issac 's orders.<br /><br />children of the corn could 've been something great , but turned into a **bad(scary)** over the top movie that you could easily make fun of . as much as i love stephen king , i 'm sure this is not what he intended and it was a pretty lame story , or at least the actors destroyed it . like i said , for a good laugh , watch it , but i 'm warning you , it 's pretty pathetic. |
| Negative | Positive | someone said that webs is a lot like an episode of sliders , and i have to agree . spoilers : i never liked the actors on sliders , and rarely have seen it except when nothing better was on . webs is the kind of movie to see if you have no other choices . read a book . webs has those kind of tv has - been actors that look like they are there as part of their probation or |

| | | work release program . some low budget tv movies have actors that at least look enthusiastic . the actors in webs look like they were getting paid minimum wage and were working on a time - clock . they have that desperate , " the - paycheck - better - not - bounce " look . the queen spider looks great , except it is rarely seen , and there are no other spiders ( and no webs ) . the queen spider bites people , and they become spider zombies , which means that they try to keep their eyes wide open when they are attacking the humans . the humans are all fighting among themselves over a number of different reasons , and they are not sympathetic . after meeting all the " humans " i would have recommended charm school for the characters . all that webs made me feel was apathy . i was numb to the characters , and hoped for some interesting gore and special effects . the gore was minimal , and the special effects were reserved for the ugly spider queen , who looked good . if webs had a bunch of spider creatures eating humans , it would have been more entertaining . apparently they could only budget " spider - zombies . " webs is a sad entry into the field of spider oriented movies . it may qualify as the **worst(worse)** spider movie ever , because eight - legged freaks had great special effects . |
|---|---|---|

**Bert Attack (SST2 Dataset):**

| Original Class | Predicted Class | Text |
|---|---|---|
| Positive | Negative | The filmmakers know how to please the **eye(camera)**, but it is not always the prettiest pictures that tell the **best(better)** story. |
| Positive | Negative | could i **have**(has) been more geeked when i heard that apollo 13 was going to **be(have)** released in imax format? |

**Bert Attack (AG News Dataset):**

| Original Class | Predicted Class | Text |
|---|---|---|
| 4 | 3 | The jury's still out on whether a computer can ever truly be **intelligent,(businessfriendly)** but there's no question that it can have multiple personalities. It's just a matter of software. |

## 8. Conclusion

An extensive analysis was conducted on how the attacks differ from each other in attack mechanism. So, as we wrap up, it was observed that our thorough investigation shed light on how different ways of attacking text vary and perform across various datasets. Key details were uncovered that can significantly assist in making natural language processing (NLP) systems more secure. In our comparison, it was revealed that PWWS Attack stands out as it focuses on making small changes to text and does so quickly. This approach is effective in situations where subtle changes are needed, outperforming methods like FBA Attack, which make bigger changes but take longer. BERT Attack falls in between, achieving a decent job at changing text and doing so at a reasonable speed without sacrificing accuracy. It was also noted that each attack method performs differently depending on the dataset used. For example, PWWS Attack performs admirably on the IMDB dataset but might not be as effective on others. Conversely, FBA Attack performs better on datasets like SST-2, highlighting the importance of understanding these differences for effective attacks. Furthermore, an examination was conducted on how well each method preserves the meaning of the original text. PWWS Attack strives to keep the meaning similar, but this might affect its accuracy on some datasets. FBA Attack, on the other hand, prioritizes accuracy over preserving meaning, indicating a balance to be struck between the two. In summary, valuable insights were provided by our study into how different text attack methods work and perform on various datasets. By understanding these differences, NLP systems can be strengthened against new threats. As the field progresses, continued exploration and innovation will be crucial for staying ahead of adversarial challenges in NLP.

## 9. Future Works

Future research in the field of adversarial attacks in multilingual natural language processing (NLP) involves gaining a comprehensive understanding of the transferability of attacks across different languages and enhancing the resilience of multilingual models against such attacks. Additionally, there's a promising avenue in exploring the impact of adversarial attacks on multimodal systems, where text is integrated with other modalities like images or audio. Moreover, there's a pressing need to devise more robust methods ensuring that adversarial examples maintain the semantic fidelity of the original text, rendering them practically indistinguishable from natural language inputs. Furthermore, researchers can delve into novel defence strategies against adversarial attacks in NLP, including adversarial training, input preprocessing techniques, and model ensemble methods, to fortify the overall robustness and security of NLP systems. Additionally, investigating adversarial attacks and defences in conversational AI systems such as chatbots or virtual assistants is crucial. This involves examining how adversarial perturbations impact the naturalness, coherence, and safety of conversational interactions, and developing defences to mitigate the adverse effects of adversarial inputs on user experience and system performance.